\documentclass[11pt,a4paper]{article}
\usepackage[hyperref]{emnlp-ijcnlp-2019}
\usepackage{times}
\usepackage{latexsym}

\usepackage{mathrsfs}
\usepackage{todonotes}
\usepackage{caption,subcaption}
\usepackage{dblfloatfix}
\usepackage[utf8]{inputenc}
\usepackage[T1]{fontenc}

\usepackage{url}
\usepackage{algorithm2e}

\aclfinalcopy 

\title{Towards Better Understanding of Spontaneous Conversations: Overcoming Automatic Speech Recognition Errors With Intent Recognition}
\author{Piotr Żelasko \\
  Avaya \\
  { \tt \small piotr.andrzej.zelasko@gmail.com }\\
  \And
  Jan Mizgajski \\
  Avaya \\
  { \tt \small mizgajski.jan@gmail.com} \\
  \AND
  Mikołaj Morzy \\
  Poznań University of Technology, \\
  Avaya \\
  { \tt \small mikolaj.morzy@put.poznan.pl} \\
  \And
  Adrian Szymczak \\
  Avaya \\
  { \tt \small adrian.dominik.szymczak@gmail.com} \\
  \AND
  Piotr Szymański \\
  Wrocław University of Science \\ and Technology, \\
  Avaya \\
  { \tt \small niedakh@gmail.com} \\
  \And
  Łukasz Augustyniak \\
  Wrocław University of Science \\ and Technology,\\
  Avaya \\
  { \tt \small luk.augustyniak@gmail.com} \\
  \AND
  Yishay Carmiel \\
  Avaya \\
  { \tt \small ycarmiel@avaya.com} \\
  }

\date{}

\begin{document}
\maketitle
\begin{abstract}
  
In this paper we present a method for correcting automatic speech recognition (ASR) errors using a finite state transducer (FST) intent recognition framework. Intent recognition is a powerful technique for dialog flow management in turn-oriented, human-machine dialogs. This technique can also be very useful in the context of human-human dialogs, though it serves a different purpose of key insight extraction from conversations. We argue that currently available intent recognition techniques are not applicable to human-human dialogs due to the complex structure of turn-taking and various disfluencies encountered in spontaneous conversations, exacerbated by speech recognition errors and scarcity of domain-specific labeled data. Without efficient key insight extraction techniques, raw human-human dialog transcripts remain significantly unexploited.
  
Our contribution consists of a novel FST for intent indexing and an algorithm for fuzzy intent search over the lattice – a compact graph encoding of ASR's hypotheses. We also develop a pruning strategy to constrain the fuzziness of the FST index search. Extracted intents represent linguistic domain knowledge and help us improve (rescore) the original transcript. We compare our method with a baseline, which uses only the most likely transcript hypothesis (best path), and find an increase in the total number of recognized intents by~25.1\%.
\end{abstract}

\section{Introduction}
\label{sec:introduction}

Spoken language understanding (SLU) consists in identifying and processing of semantically meaningful parts of the dialogue, most often performed on the transcript of the dialogue produced by the automatic speech recognition (ASR) system~\cite{ward1991understanding}.  These meaningful parts are referred to as \emph{dialog acts} and provide structure to the flow of conversation. Examples of dialog acts include statements, opinions, yes-no questions, backchannel utterances, response acknowledgements, etc~\cite{Stolcke2000}. The recognition and classification of dialog acts is not sufficient for true spoken language understanding. Each dialog act can be instantiated to form an \emph{intent}, which is an expression of a particular intention. Intents are simply sets of utterances which exemplify the intention of the speaking party to perform a certain action, convey or obtain information, express opinion, etc. For instance, the intent \emph{Account Check} can be expressed using examples such as "\emph{let me go over your account}", "\emph{found account under your name}", "\emph{i have found your account}". At the same time, the intent \emph{Account Check} would be an instance of the dialog act \emph{Statement}. An important part of intent classification is entity recognition~\cite{nadeau2007survey}. An \emph{entity} is a token which can either be labeled with a proper name, or assigned to a category with well defined semantics. The former leads to the concept of a \emph{named entity}, where a token represents some real world object, such as a location (\emph{New York}), a person (\emph{Lionel Messi}), a brand (\emph{Apple}). The latter can represent concepts such as money (\emph{one hundred dollars}), date (\emph{first of May}), duration (\emph{two hours}), credit card number, etc.

\subsection{Motivation}

Spoken language understanding is a challenging and difficult task~\cite{Furui2002}. All problems present in natural language understanding are significantly exacerbated by several factors related to the characteristics of spoken language. 

Firstly, each ASR engine introduces a mixture of systematic and stochastic errors which are intrinsic to the procedure of transcription of spoken audio. The quality of transcription, as measured by the popular word error rate (WER), attains the level of 5\%-15\% WER for high quality ASR systems for English~\cite{povey2016purely, han2017capio, xiong2018microsoft, hahm2018marchex}. The WER highly depends on the evaluation data difficulty and the speed to accuracy ratio. Importantly, errors in the transcription appear stochastically, both in audio segments which carry important semantic information, as well as in inessential parts of the conversation.

Another challenge stems from the fact that the structure of conversation changes dramatically when a human assumes an agency in the other party. When humans are aware that the other party is a machine (as is the case in dialogue chatbot interfaces), they tend to speak in short, well-structured turns following the subject-verb-object (SVO) sentence structure with a minimal use of relative clauses~\cite{hill2015real}. This structure is virtually nonexistent in spontaneous speech, where speakers allow for a non-linear flow of conversation. This flow is further obscured by constant interruptions from backchannel or cross-talk utterances, repetitions, non-verbal signaling, phatic expressions, linguistic and non-linguistic fillers, restarts, and ungrammatical constructions. 

A substantial, yet often neglected difficulty, stems from the fact that most SLU tasks are applied to transcript segments representing a single turn in a typical scripted conversation. However, turn-taking in unscripted human-human conversations is far more haphazard and spontaneous than in scripted dialogues or human-bot conversations. As a result, a single logical turn may span multiple ASR segments and be interwoven with micro-turns of the other party or the contrary - be part of a larger segment containing many logical turns. 

ASR transcripts lack punctuation, normalization, true-casing of words, and proper segmentation into phrases as these features are not present in the conversational input~\cite{zelasko2018punctuation}. These are difficult to correct as the majority of NLP algorithms have been trained and evaluated on text and not on the output of an ASR system. Thus, a simple application of vanilla NLP tools to ASR transcripts seldom produces actionable and useful results. 

Finally, speech based interfaces are defined by a set of dimensions, such as domain and vocabulary (retail, finance, entertainment), language (English, Spanish), application (voice search, personal assistant, information retrieval), and environment (mobile, car, home, distant speech recognition). These dimensions make it very challenging to provide a low cost domain adaptation. 

Last but not least, production ASR systems impose strict constraints on the additional computation that can be performed. Since we operate in a near real-time environment, this precludes the use of computationally expensive language models which could compensate for some of the ASR errors.

\subsection{Contribution}

We identify the following as the key contributions of this paper:

\textbf{A discussion of intent recognition in human-human conversations.} While significant effort is being directed into human-machine conversation research, most of it is not directly applicable to human-human conversations. We highlight the issues frequently encountered in NLP applications dealing with the latter, and propose a framework for intent recognition aimed to address such problems.

\textbf{A novel FST intent index construction with dedicated pruning algorithm, which allows fuzzy intent matching on lattices.} To the best of our knowledge, this is the first work offering an algorithm which performs a fuzzy search of intent phrases in an ASR lattice, as opposed to a linear string. We build on the well-studied FST framework, using composition and sigma-matchers to enable fuzzy matching, and extend it with our own pruning algorithm to make the fuzzy matching behavior correct. We supply the method with several heuristics to select the new best path through the lattice and we confirm their usefulness empirically. Finally, we ensure that the algorithm is efficient and can be used in a real-time processing regime.

\textbf{Domain-adaptation of an ASR system in spite of data scarcity issues.} Generic ASR systems tend to be lackluster when confronted with specialized jargon, often very specific to a single domain (e.g., healthcare services). Creating a new ASR model for each domain is often impractical due to limited in-domain data availability or long training times. Our method improves the speech recognition, without the need for any re-training, by improving the recognition recall of the anticipated intents – the key insight sources in these conversations.

\section{Related Work}
\label{sec:related_work}

\subsection{Domain Knowledge Modeling for Machine Learning}
\label{sec:human-knowl}

The power of some of the best conversational assistants lies in domain-dependent human knowledge. 
Amazon's Alexa is improving with the user generated data it gathers  \cite{Kumar}. 
Some of the most common human knowledge base structures used in NLP are word lists such as dictionaries for ASR \cite{Bach2007HandlingOW}, sentiment lexicons \cite{e18010004} knowledge graphs such as WordNet \cite{MazPiaRudSzpaKedz:16, Miller95wordnet:a} and ConceptNet \cite{L12-1639}. 
Conceptually, our work is similar to~\citet{47597}, however, they do not allow for fuzzy search through the lattice.

\subsection{Word Confusion Networks}

\begin{figure*}[ht!]
    \centering
    \includegraphics[width=\textwidth,height=\textheight,keepaspectratio]{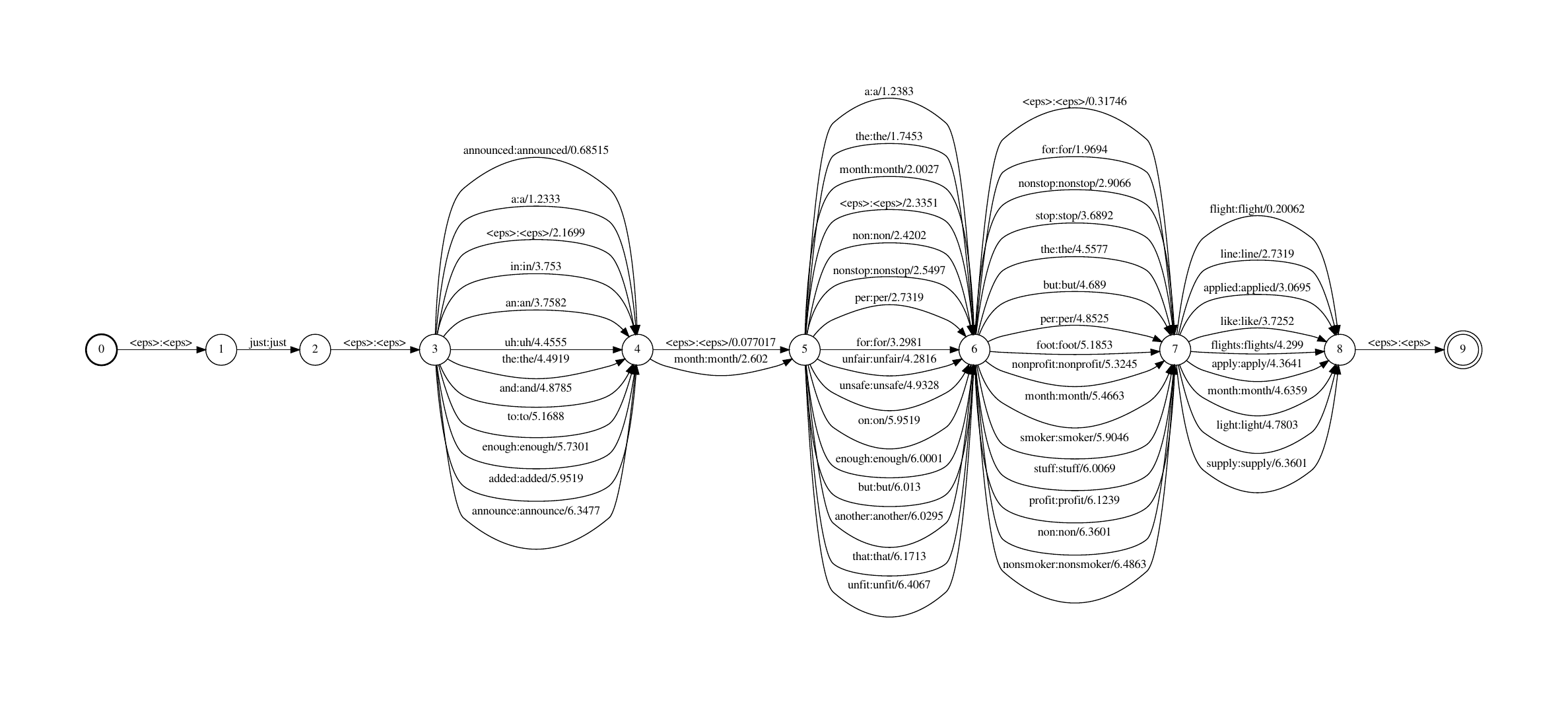}
    \vspace{-15mm}
    \caption{Word confusion network for the utterance \textit{just a nonstop flight}. $\left\langle \mathit{eps} 
    \right\rangle$ indicates an empty transition.}
    \label{fig:wcn}
\end{figure*}

A word confusion network (WCN) \cite{mangu2000finding} is a highly compact graph representation of possible confusion cases between words in the ASR lattice. The nodes in the network represent words and are weighted by the word's confidence score or its \emph{a posteriori} probability. Two nodes (words) are connected when they appear in close time points and share a similar pronunciation, which merits suspecting they might get confused in recognition \cite{stolcke2002srilm,hakkani2003general}. WCN may contain empty transitions which introduce paths through the graph that skip a particular word and its alternatives. 
An example of WCN is presented in Figure~\ref{fig:wcn}. Note that this seemingly small lattice encodes 46~080 possible transcription variants.

Various language understanding tasks have been improved in recent years using WCNs: language model learning \cite{gretter2001line}, ASR improvements \cite{tur2002improving,HAKKANITUR2006495,6289024}, classification \cite{cortes2003lattice,masumura2018neural}, word spotting  \cite{hori2007open,zhang2007keyword}, voice search \cite{feng2009effects}, dialog state tracking \cite{jagfeld2017encoding} and named entity extraction \cite{HAKKANITUR2006495,KURATA2012491,hakkani2014probabilistic}. \citet{stiefel2017enriching} modified the WCN approach to include part-of-speech information in order to achieve an improvement in semantic quality of recognized speech.

\subsection{Finite State Transducers}
The finite state transducer (FST) \cite{roche1997finite,mohri2004weighted} is a finite state machine with two memory tapes that maps input symbols to output symbols as it reads from the input table and writes to the output tape. FSTs are natural building blocks for systems that transform one set of symbols into another due to the robustness of various FST joining operations such as union, concatenation or composition. Composing FST1 and FST2 is performed by running an input through the FST1, taking its output tape as the input tape for FST2 and returning the output of FST2 as the output of the composed FST. For a formal definition of the operation and a well-illustrated example, we refer the reader to \cite{argueta2018composing}. 

Finite state transducers have been widely used in speech recognition \cite{lehr2011learning,mohri2002weighted, moore2006juicer}, named entity recognition \cite{friburger2004finite, gaio2017extended}, morpho-syntactic tagging \cite{roche1995deterministic,forsberg2016learning,moeller2018neural} or language generation \cite{goyal2016natural}.

\section{Methods}
\label{sec:methods}

\subsection{Automatic Speech Recognition}
\label{sec:methods:asr}

To transcribe the conversations we use an ASR system built using the Kaldi toolkit~\cite{povey2011kaldi} with a TDNN-LSTM acoustic model trained with lattice-free maximum mutual information (LF-MMI) criterion~\cite{povey2016purely} and a 3-gram language model for utterance decoding. The ASR lattice is converted to a word confusion network (WCN) using minimum Bayes risk (MBR) decoding~\cite{xu2011minimum}.

\subsection{Domain Knowledge Acquisition - Intent Definition and Discovery}
\label{sec:methods:intent_definition}

While an in-depth description of tools used in the intent definition process is beyond the scope of this paper, we provide a brief overview to underline the application potential of our algorithm when combined with a sufficient body of domain knowledge. First, let us formalize the notion of intents and intent examples. An intent example is a sequence of words which conveys a particular meaning, e.g., "\emph{let me go over your account}" or "\emph{this is outrageous}". An intent is a collection of intent examples conveying a similar meaning, which can be labeled with an intelligible and short description helpful in understanding the conversation.

Some of the intents that we find useful include customer requests (Refund, Purchase Intent), desired actions by the agent (Up-selling, Order Confirmation) or compliance and customer satisfaction risks (Customer Service Complaint, Supervisor Escalation).
Defining all examples by hand would be prohibitively expensive and cause intents to have limited recall and precision, as, by virtue of combinatorial complexity of language, each intent needs hundreds of examples. To alleviate this problem we provide annotators with a set of tools, including: fast transcript annotation user interface for initial discovery of intents; an interactive system for semi-automatic generation of examples which recommends synonyms and matches examples on existing transcripts for validation; unsupervised and semi-supervised methods based on sentence similarity and grammatical pattern search for new intent and examples discovery.

In addition, we extend the notion of an example with two concepts that improve the recall of a single example:

\begin{itemize}
    \item \emph{Blank quota}, that defines the number of words that may be found \emph{in-between} the words of the example and still be acceptable, e.g., "\emph{this is \emph{very} outrageous}" becomes a potential match for "\emph{this is outrageous}" if the blank quota is greater than 0. This allows the annotator to focus on the words that convey the meaning of the phrase and ignore potential filler words.
    \item \emph{Entity templating} allowing examples to incorporate entities in their definitions. With entity templating an example "\emph{your flight departs \emph{\_\_SYSTEM\_TIME\_\_}}" would match both "\emph{your flight departs \emph{in ten minutes}}" and "\emph{your flight departs \emph{tomorrow at seven forty five p m}}". This relieves the annotator from enumerating millions of possible examples for each entity and facilitates the creation of more specific examples that increase precision. To illustrate, "\emph{your item number is}" could incorrectly match "\emph{your item number is wrong}", but "\emph{your item number is \emph{\_\_SYSTEM\_NUMBER\_\_}}" would not. 
\end{itemize}

The above methods allow the annotators to create hundreds of intents efficiently, with thousands of examples, allowing millions of \emph{distinct} potential phrases to be matched. When combined with the ability for customers to configure entities and select a subset of intents that are relevant to their business, this approach produces highly customer-specific repositories of domain knowledge.

\subsection{Lattice Rescoring Algorithm}
\label{sec:methods:lattice_rescoring}

\begin{figure}[htb]
    \centering
    \includegraphics[width=\textwidth/2,height=\textheight,keepaspectratio]{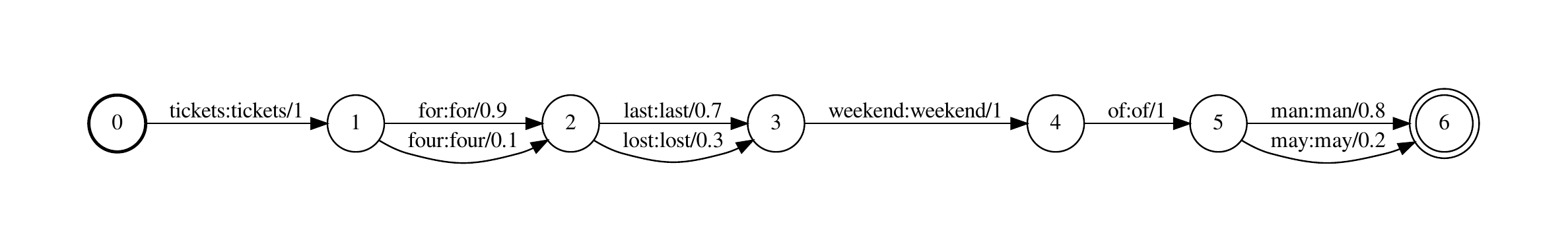}
    \vspace{-10mm}
    \caption{Word confusion network representing the lattice $\mathcal{L}$.}
    \label{fig:wcn_for_composition}
\end{figure}

The lattice $\mathcal{L}$ is an acceptor, where each arc contains a symbol representing a single word in the current hypothesis (see Figure~\ref{fig:wcn_for_composition}). We employ a closed ASR vocabulary assumption and operate on word-level, rather than character- or phoneme- level FST. Note that this assumption is not a limitation of our method. Should the ASR have an unlimited vocabulary (as some end-to-end ASR systems do), it is possible to dynamically construct the lattice symbol table and merge it with the symbol table of intent definitions.  

To perform intent annotation (i.e., to recognize and mark the position of intent instances in the transcript), we first create the FST index $\mathcal{I}$ of all intent examples. This index is a transducer which maps the alphabet of words (input symbols) onto the alphabet of intents (output symbols). We construct index $\mathcal{I}$ in such a way that its composition with the lattice results in another transducer $\mathcal{A} = \mathcal{L} \circ \mathcal{I}$ representing the annotated lattice. 

\begin{figure}[htb]
    \centering
    \includegraphics[width=\textwidth/2,height=\textheight,keepaspectratio]{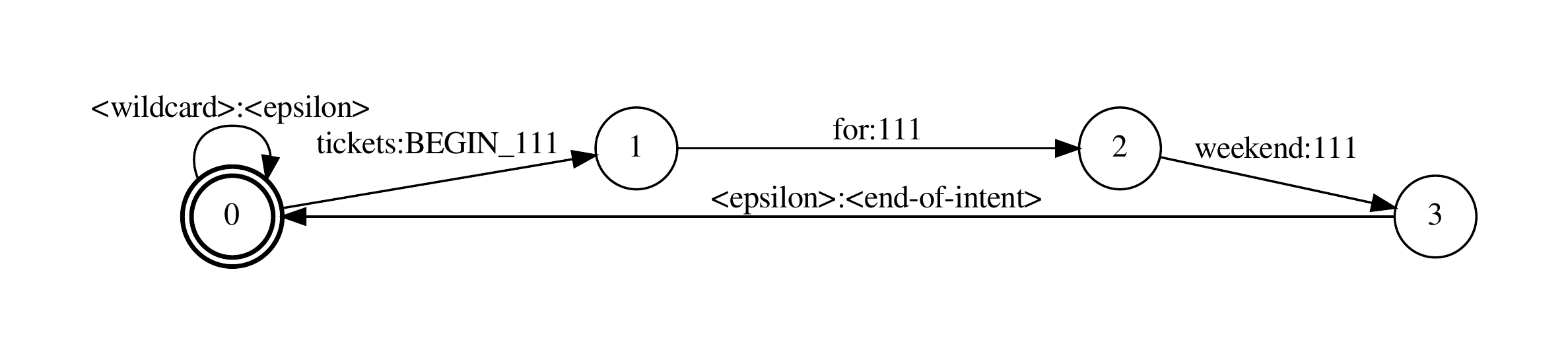}
    \vspace{-10mm}
    \caption{An index $\mathcal{I}$ matching a single intent example \textit{tickets for weekend} to an intent number 111.}
    \label{fig:trivial_index}
\end{figure}

We begin by creating a single FST state which serves as both the initial and the final state and contains a single loop wildcard arc. A wildcard arc accepts any input symbol and transduces it to an empty $\epsilon$ output symbol. The wildcard arc can be efficiently implemented with special $\sigma$-matchers, available in the OpenFST framework~\cite{allauzen2007openfst}.
Composition with the singleton FST maps every input symbol in $\mathcal{L}$ to $\epsilon$, which denotes the lack of intent annotations. For each intent example, we construct an additional branch (i.e. a set of paths) in the index $\mathcal{I}$ which maps multiple sequences of words to a set of symbols representing this particular intent example (see Figure~\ref{fig:trivial_index}). 

We use three types of symbols: an intent symbol $\iota$ (including begin $\iota_B$, continuation $\iota_C$ and end $\iota_E$ symbols), an entity symbol $\omega$ (including an entity placeholder symbol $\omega^*$), and a null symbol $\epsilon$.

The intent symbol $\iota$ is the delimiter of the intent annotation and it demarcates the words constituting the intent. The begin ($\iota_B$) and continuation ($\iota_C$) symbols are mapped onto arcs with words as input symbols, and the end ($\iota_E$) symbol is inserted in an additional arc with an $\epsilon$ input symbol after the annotated fragment of text. It is important that the begin symbol $\iota_B$ does not introduce an extra input of $\epsilon$. Otherwise, the FST composition is inefficient, as it tries to enter this path on every arc in the lattice $\mathcal{L}$.

\begin{figure}[h]
    \centering
    \includegraphics[width=\textwidth/2,height=\textheight/2,keepaspectratio]{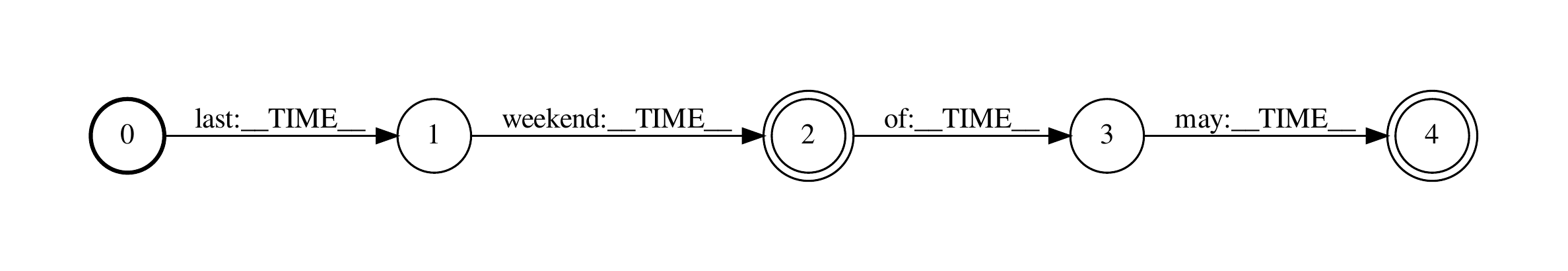}
    \vspace{-10mm}
    \caption{A simple grammar FST $\mathcal{E}$ for the non-terminal token \textit{\_\_TIME\_\_}. Note that both states 2 and 4 are final (indicated by the double circle).}
    \label{fig:time_grammar}
\end{figure}

The entity symbol $\omega$ marks the presence of an entity in the intent annotation. Each entity in the intent index $\mathcal{I}$ is constructed as a non-terminal entity placeholder $\omega^*$, which allows using the FST lazy replacement algorithm to enter a separate FST grammar $\mathscr{E}$ describing a set of possible values for a given entity. We use the $\mathscr{E}$ transducer when it is possible to provide a comprehensive list of entity instances. Otherwise, we provide an approximation of this list by running a named entity recognition model predictions on an n-best list (see Figure~\ref{fig:time_grammar}). Finally, the null symbol $\epsilon$ means that either no intent was recognized, or the word spanned by the intent annotation did not contribute to the annotation itself.

\begin{figure*}[htb]
    \centering
    \includegraphics[width=\textwidth,height=\textheight,keepaspectratio]{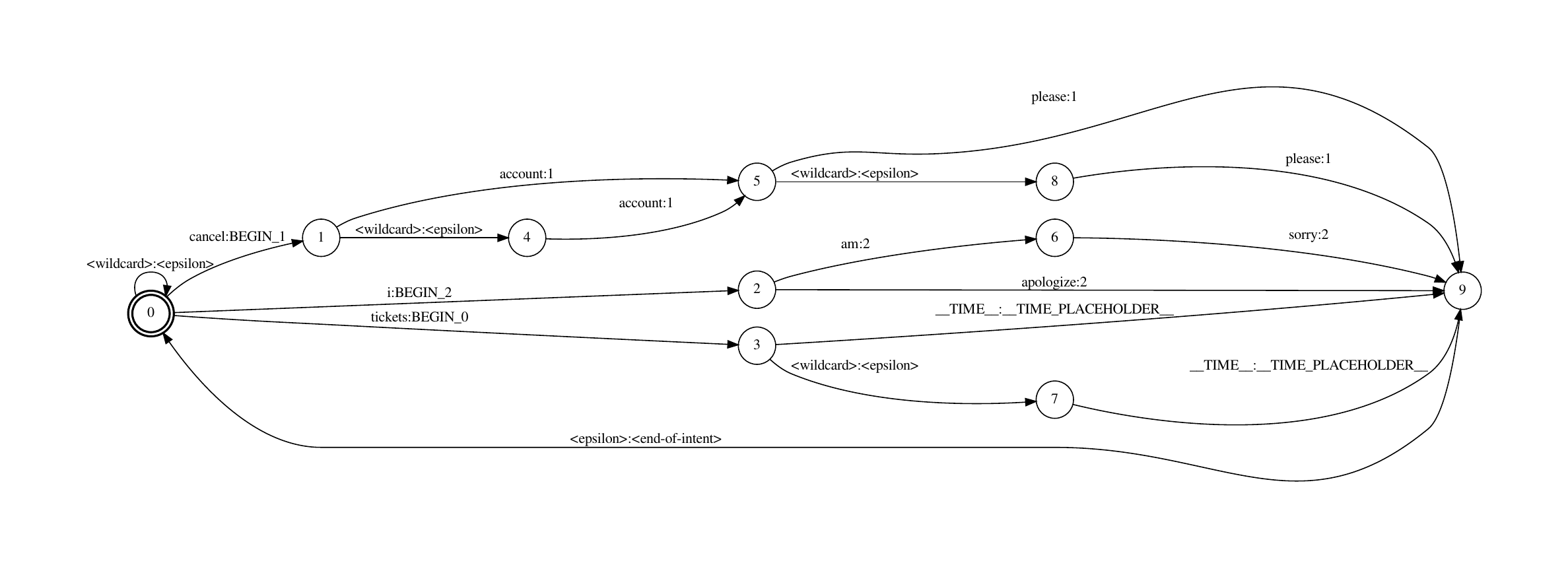}
    \vspace{-15mm}
    \caption{The intent index $\mathcal{I}$ which matches three different intent examples: \textit{cancel account please} with a blank quota of 1, \textit{i apologize} with the synonym \textit{am sorry}, and \textit{tickets \_\_SYSTEM\_TIME\_\_}, where the last token is a special non-terminal token, replaced dynamically during composition with an appropriate grammar FST.}
    \label{fig:entity_index}
\end{figure*}
\begin{figure*}[htb]
    \centering
    \begin{subfigure}[b]{\linewidth}
        \includegraphics[width=\textwidth,height=\textheight,keepaspectratio]{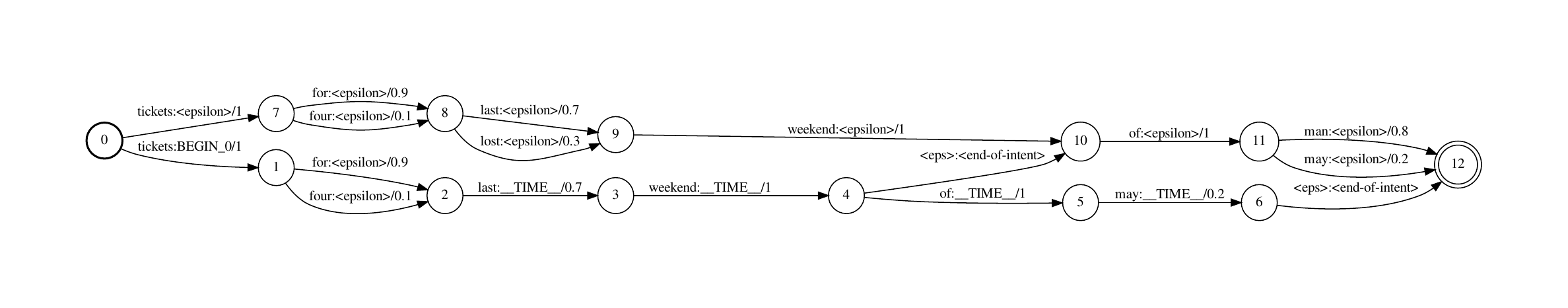}
        \vspace{-15mm}
        \caption{\label{fig:before_pruning}}
    \end{subfigure}
    \begin{subfigure}[b]{\linewidth}
        \includegraphics[width=\textwidth,height=\textheight,keepaspectratio]{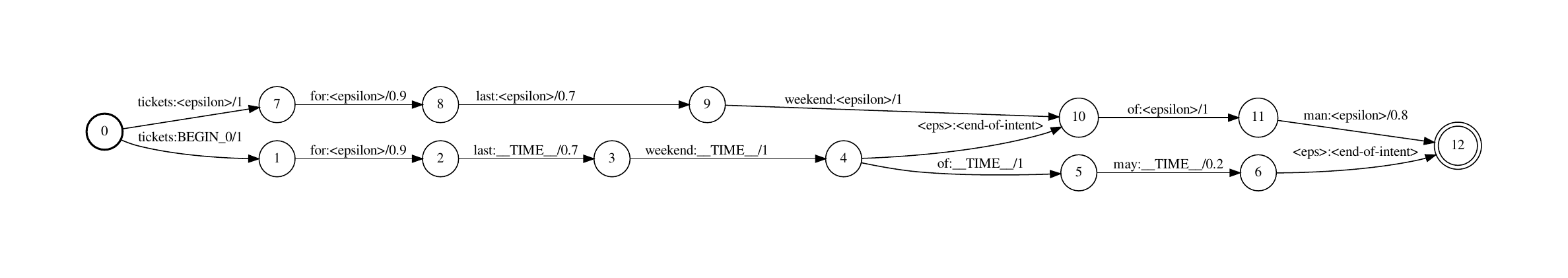}
        \vspace{-15mm}
        \caption{\label{fig:after_pruning}}
    \end{subfigure}
    \caption{Annotated lattice $\mathcal{A}$ resulting from composition $\mathcal{L'} \circ \mathcal{I}$ with replacement using $\mathscr{E}$ before (a) and after (b) pruning. Note that the last word \textit{man} was rescored as \textit{may} in the path due to the recognition of an annotated intent.}
    \label{fig:annotated_lattice}
\end{figure*}

This procedure successfully performs exact matching of the transcription to intent index, when all words present in the current lattice path are also found in the intent example. Unfortunately, this approach is highly impractical when real transcriptions are considered. Sequences of significant words are interwoven with filler phonemes and words, for instance the utterance "\emph{I want uhm to order like um three yyh three tickets}" could not be matched with the intent example "\emph{I want to order \_\_NUMBER\_\_ tickets}". 

To overcome this limitation we adapt the intent index $\mathcal{I}$ to enable fuzzy matching so that some number $n$ of filler words can be inserted between words constituting an intent example, while still allowing to match the intent annotation. We add wildcard arcs between each of the intent-matching words, to provide the matching capacity of $0$ to $n$ matches of any word in the alphabet. The example of such an index is shown in Figure~\ref{fig:entity_index}. 

The naive implementation allowing for $n$ superfluous (non-intent) words appearing between intent-matching words would lead to a significant explosion of the annotations spans. Instead, we employ a post-processing filtering step to prune lattice paths where the number of allowed non-intent word is exceeded. Our filtering step has a computational complexity of $\mathcal{O}(|S| + |A|)$, where $|S|$ is the number of states and $|A|$ is the number of arcs in the non-pruned annotated lattice $\mathcal{L'}$. 

The pruning algorithm is based on the depth-first search (DFS) traversal of the lattice $\mathcal{L'}$ and marks each state in $\mathcal{L'}$ as either new, visited, or pruned. Only new states are entered and each state is entered at most once. The FST arcs are only marked as either visited or pruned. Each FST state keeps track of whether an intent annotation parsing has begun (i.e., a begin symbol $\iota_B$ has been encountered but the end symbol $\iota_E$ has not appeared yet) and how many wildcard words have been matched so far. 

The traversal of the lattice $\mathcal{L'}$ is stateful. It starts in a \textit{non-matching} state and remains in this state until encountering the intent begin symbol $\iota_B$. Then the state becomes \textit{matching} and remains such until encountering the intent end symbol $\iota_E$. A state is marked as \textit{pruned} when the traversal state is \textit{matching} and the number of wildcard words exceeds the blank quota for the given intent example. Any arc incident with a pruned state is not entered during further traversal, leading to a significant speed-up of lattice processing. After every possible path in the lattice $\mathcal{L'}$ has been either traversed or pruned, all redundant (i.e., pruned or not visited) FST states are removed from the lattice, along with all incident arcs.

\begin{table*}[!t]
\begin{center}
\begin{tabular}{|p{4.3cm}|p{4.7cm}|p{5.5cm}|}
\hline \textbf{Intent} &  \textbf{Original text} &    \textbf{Rescored text} \\ \hline
                Website Mention (with entity \textit{Brand}) &   just use your regular acme (ok) that calm pay with credit card & just use your regular acme (ok) dot com pay with credit card \\ \hline
                Question: Account Lookup &  can you looked at my count &  can you look at my account \\ \hline
                Question: Account Information &  i need to recount (mhm sure) number or email addresses & i need your account (mhm sure) number or email address \\ \hline
                Call Opening &   think of a collie level & thank you for calling \\ \hline
                End of Hold &   thank you for your patients & thank you for your patience \\ \hline
                Refund &   work connie the refined & work on the refund \\
\hline
\end{tabular}
\end{center}
\vspace{-3mm}
\caption{Examples of successful lattice rescoring along with the recognized intent. In first example real brand name was obfuscated by a fictional brand name ACME. Words in parentheses are turns of another speaker.}
\label{tab:rescoring_examples}
\end{table*}

The annotated lattice $\mathcal{A}$ is obtained after final traversal of the lattice $\mathcal{L'}$ which prunes arcs representing unmatched word alternatives. If no intent has been matched on any of the parallel arcs, the traversal retains only the best path hypothesis. Figure~\ref{fig:annotated_lattice} presents an example of the annotated lattice before and after pruning.

\subsection{Parsing the Annotated Lattice}
\label{sec:methods:parsing_annotated_lattice}

Despite significant pruning described in the previous section, the annotated lattice $\mathcal{A}$ still contains competing variants of the transcript. The next step consists in selecting the "best" variant and traversing all paths in $\mathcal{A}$ which correspond to this transcript. The key concept of our method is to guide the selection of the "best" variant by intents rather than word probabilities. We observe that the likelihood of a particular longer sequence of words in the language is lower than the likelihood of a particular shorter sequence of words. Since \emph{a priori} longer intent examples are less likely to appear in the lattice $\mathcal{A}$ purely by chance, the presence of a lattice path containing a longer intent example provides strong evidence for that path. 

The complete set of heuristics applied sequentially to the annotated lattice $\mathcal{A}$ in search of the best path is the following: (a) select the path with the longest intent annotation; (b) select the path with the largest number of intent annotations; (c) select the path with the intent annotation with the longest span (i.e. consider also blank words), (d) select the path with the highest original ASR likelihood. The chosen best path is composed with the annotated lattice $\mathcal{A}$ to produce the annotated lattice $\mathcal{A^*}$ with the final variant of the transcript. The output intent annotations are retrieved by traversing every path in $\mathcal{A^*}$.

\subsection{Lattice concatenation}
\label{sec:methods:lattice_concatenation}

As hinted in Section~\ref{sec:introduction}, most NLP tasks are performed on the turn level, which naturally corresponds to the turn-taking patterns in a typical human-machine dialogue. This approach yields good results for chatbot conversational interfaces or information retrieval systems, but for spontaneous human-human dialogues, the demarcation of turns is much more difficult due to the presence of fillers, interjections, ellipsis, backchannels, etc. Thus, we cannot expect those intent examples would align with ASR segments which capture a single speaker turn. We address this issue by concatenating turn-level lattices of all utterances of a person throughout the conversation into a conversation-lattice $\mathcal{L^C}$. This lattice can still be effectively annotated and pruned using algorithms presented in Section~\ref{sec:methods:lattice_rescoring} to obtain the annotated conversation-lattice $\mathcal{A^C}$.

Unfortunately, the annotated conversation-lattice $\mathcal{A^C}$ cannot be parsed in search of the best path using the algorithm presented in Section~\ref{sec:methods:parsing_annotated_lattice}, because the computational cost of every path traversal in $\mathcal{A^C}$ is exponential in the number of words. Fortunately, we can exploit the structure of the conversation-lattice $\mathcal{A^C}$ to identify the best path. We observe that $\mathcal{A^C}$ is a sequence of segments organized either in series or in parallel. Segments with no intent annotations are series of linear word hypotheses, which branch to parallel word hypotheses whenever an intent annotation is matched (because the original path with no intent annotation is retained in the lattice). The parallel segment ends with the end of the intent annotation. These series and parallel segments can be detected by inspecting the cumulative sum of the difference of out-degree and in-degree of each state in a topologically sorted conversation-lattice $\mathcal{A^C}$.  For series regions, this sum will be equal to 0, and greater than 0 in parallel regions. The computational cost of performing this segmentation is $\mathcal{O}(|S| + |A|)$, i.e., linear in the number of states and arcs in the annotated conversation-lattice $\mathcal{A^C}$. After having performed the segmentation, the partial best path search in parallel segments is resolved using the method presented in Section~\ref{sec:methods:parsing_annotated_lattice}.

\section{Experimental results}
\label{sec:analysis}

In this section, we present a quantitative analysis of the proposed algorithm. 
The baseline algorithm annotates only the best ASR hypothesis.
We perform the experiments with an intent library comprised of 313 intents in total, each of which is expressed using 169 examples on average. 
The annotations are performed on more than 70~000 US English phone conversations with an average duration of 11 minutes, but some of them take even over one hour.
The topics of these conversations span across several domains, such as inquiry for account information or instructions, refund requests or service cancellations.
Each domain uses a relevant subset of the intent library (typically 100-150 intents are active).

To evaluate the effectiveness of the proposed algorithm, we have sampled a dataset of 500 rescored intent annotations found in the lattices in cancellations and refunds domain. The correctness of the rescoring was judged by two annotators, who labeled 250 examples each. The annotators read the whole conversation transcript and listened to the recording to establish whether the rescoring is meaningful. In cases when a rescored word was technically incorrect (e.g., mistaken tense of a verb), but the rescoring led to the recognition of the correct intent, we labeled the intent annotation as correct. The results are shown in Table~\ref{tab:accuracy_per_intent_length}. Please note that every result above 50\% indicates an improvement over the ASR best path recognition, since we correct more ASR errors than we introduce new mistakes.

\begin{table}[!h]
\begin{center}
\begin{tabular}{|c|c|c|}
\hline \textbf{Intent length} &  \textbf{Occurrences} &    \textbf{Accuracy [\%]} \\ \hline
                1 &   25 &   32.0 \\
                2 &  139 &   39.5 \\
                3 &  149 &   63.7 \\
                4 &   80 &   76.2 \\
                5 &   53 &   94.3 \\
                6 &   19 &   94.7 \\
               7+ &   35 &  100.0 \\
\hline
\end{tabular}
\end{center}
\vspace{-3mm}
\caption{Rescoring accuracy w.r.t.~intent length measured on an annotated dataset of 500 rescored intents.}
\label{tab:accuracy_per_intent_length}
\end{table}

The results confirm our assumptions presented in Section~\ref{sec:methods:parsing_annotated_lattice}. The longer the intent annotation, the more likely it is to be correct due to stronger contextuality of the annotation. Intent annotations which span at least three words are more likely to rescore the lattice correctly than to introduce a false positive. These results also lead us to a practical heuristic, that an intent annotation which spans only one or two words should not be considered for rescoring. Application of this heuristic results in an estimated accuracy of 77\%. We use this heuristic in further experiments. A stricter heuristic would require at least four words span, with an accuracy of 87.7\%. Calibration of this threshold is helpful when the algorithm is adapted to a downstream task, where a different precision/recall ratio may be required. We present some examples of successful lattice rescoring in Table~\ref{tab:rescoring_examples}.

The proposed algorithm finds 658~549 intents in all conversations, covering 4.1\% of all (62~450~768) words, whereas the baseline algorithm finds 526 356 intents, covering 3.3\% of all words. Therefore, the increase in intent recognition of the method is 25.1\% by rescoring 8.3\% of all annotated words (0.34\% of all words). Particular intents achieve different improvements ranging from no improvement up to 1062\% -- ranked percentile results are presented in Table~\ref{tab:percetile_improvement}. We see that half of intents gain at least 35.7\% of improvement, while 20\% of all intents gain at least 83.5\%.

\begin{table}[!h]
\begin{center}
\begin{tabular}{|c|c|}
\hline \textbf{Intent classes [\%]} &    \textbf{Min. improvement [\%]} \\ \hline
                10 &   128.9 \\
                20 &   83.5 \\
                30 &   62.4 \\
                40 &   49.4 \\
                50 &   35.7 \\
                60 &   28.7 \\
                70 &   21.6 \\
                80 &   13.7 \\
                90 &   2.0 \\
                
\hline
\end{tabular}
\end{center}
\vspace{-3mm}
\caption{Ranked percentiles of improvement in intent recognition. The improvement is determined for each intent class individually. Intent classes are sorted and binned into percentiles, for each bin we report the minimum improvement for intents in the bin.}
\label{tab:percetile_improvement}
\end{table}

\section{Conclusions}
\label{sec:conclusions}

A commonly known limitation of the current ASR systems is their inability to recognize long sequences of words precisely. In this paper, we propose a new method of incorporating domain knowledge into automatic speech recognition which alleviates this weakness. Our approach allows performing fast ASR domain adaptation by providing a library of intent examples used for lattice rescoring. The method guides the best lattice path selection process by increasing the probability of intent recognition. At the same time, the method does not rescore paths of unessential turns which do not contain intent examples. As a result, our approach improves the understanding of spontaneous conversations by recognizing semantically important transcription segments while adding minimal computational overhead. Our method is domain agnostic and can be easily adapted to a new one by providing the library of intent examples expected to appear in the new domain. The increased intent annotation coverage allows us to train more sophisticated models for downstream tasks, opening the prospects of true spoken language understanding.


\bibliography{emnlp-ijcnlp-2019}
\bibliographystyle{acl_natbib}

\appendix


\end{document}